\documentclass{article}

\usepackage[preprint]{neurips}

\usepackage{natbib}
\setcitestyle{numbers,square}

\usepackage[utf8]{inputenc} 
\usepackage[T1]{fontenc}    
\usepackage{hyperref}       
\usepackage{url}            
\usepackage{booktabs}       
\usepackage{amsfonts}       
\usepackage{nicefrac}       
\usepackage{microtype}      
\usepackage{xcolor}         
\usepackage{subcaption}

\usepackage{amsmath}
\usepackage{xspace}
\usepackage{amssymb}
\usepackage{graphicx}
\usepackage{multirow}
\usepackage{enumitem}
\usepackage{ulem}
\usepackage{float}
\usepackage{makecell}
\usepackage{caption}
\usepackage{threeparttable}
\usepackage{fancyhdr}
\newcommand{\eg}{\textit{e.g.,}\xspace}

\newcommand{\paratitle}[1]{\vspace{0.8ex}\noindent \textbf{#1}}

\newcommand{\modelname}{Tele-FLM\xspace}

\title{52B to 1T: Lessons Learned via Tele-FLM Series}

\author{
  Xiang Li\textsuperscript{1\textdagger}, 
  Yiqun Yao\textsuperscript{1\textdagger},
  Xin Jiang\textsuperscript{1\textdagger}, 
  Xuezhi Fang\textsuperscript{1\textdagger}, 
  Chao Wang\textsuperscript{2\textdagger},
  Xinzhang Liu\textsuperscript{2\textdagger},\\
  \textbf{
  Zihan Wang\textsuperscript{2},
  Yu Zhao\textsuperscript{2}, 
  Xin Wang\textsuperscript{2},
  Yuyao Huang\textsuperscript{2}, 
  Shuangyong Song\textsuperscript{2},
  Yongxiang Li\textsuperscript{2},
  }\\
  \textbf{
  Zheng Zhang\textsuperscript{1}, 
  Bo Zhao\textsuperscript{1}, 
  Aixin Sun\textsuperscript{3}, 
  Yequan Wang\textsuperscript{1$*$},
  Zhongjiang He\textsuperscript{2$*$},
  } \\
  \textbf{
  Zhongyuan Wang\textsuperscript{1},
  Xuelong Li\textsuperscript{2},
  Tiejun Huang\textsuperscript{1}
  }\\ \\
  $^{1}$Beijing Academy of Artificial Intelligence, Beijing, China\\
  $^{2}$Institute of Artificial Intelligence (TeleAI), China Telecom Corp Ltd, China\\
  $^{3}$School of Computer Science and Engineering, Nanyang Technological University, Singapore
}
\date{June 2024}

\begin{document}

\maketitle
\renewcommand{\thefootnote}{\fnsymbol{footnote}}
\footnotetext[2]{Indicates equal contribution.}
\footnotetext[1]{Corresponding authors.}
\renewcommand{\thefootnote}{\arabic{footnote}}

\begin{abstract}
Large Language Models (LLMs) represent a significant stride toward Artificial General Intelligence. As scaling laws underscore the potential of increasing model sizes, the academic community has intensified its investigations into LLMs with capacities exceeding 50 billion parameters. This technical report builds on our prior work with \modelname (also known as FLM-2), a publicly available 52-billion-parameter model. We delve into two primary areas: we first discuss our observation of Supervised Fine-tuning (SFT) on Tele-FLM-52B, which supports the ``less is more'' approach for SFT data construction; second, we demonstrate our experiments and analyses on the best practices for progressively growing a model from 52 billion to 102 billion, and subsequently to 1 trillion parameters. We will open-source a 1T model checkpoint, namely Tele-FLM-1T, to advance further training and research.

\end{abstract}

\section{Introduction}
Large language models (LLMs) \cite{vaswani2017attention, radford2019language, gemini, GPT-4} have demonstrated remarkable general capabilities \cite{sparks}. Research on scaling laws \cite{scaling-law, scaling-law-2, chinchilla, mu-scaling} indicates that metrics related to perplexity (\eg loss and BPB) improve as training FLOPs increase. Given that high-quality data may be limited due to factors such as copyright constraints and the proliferation of LLM-generated content on the web, there is a growing interest within the community in scaling up model sizes. Recent iterations of popular LLM series, such as Mistral at 141B \cite{mistral}, DeepSeek at 236B \cite{deepseek}, Grok at 314B \cite{grok}, and Llama-3 exceeding 400B parameters \cite{llama3}, underscore a trend toward models with 1 trillion parameters. To benefit the explorations on extremely large language models, we trained a series of models, namely Tele-FLM (a.k.a. FLM-2), in which we first train a 52B model, and grow it to 1T parameters, with an intermediate stage of 102B. We outline techniques for efficiently and robustly training the 52B model in \cite{teleflm}. As a consequent work, we focus on two prominent areas of research with the Tele-FLM models: \textit{alignment with human} \cite{instructgpt,alpaca} and \textit{progressive learning} \cite{bert2bert,msg,flm101b}.

To align with humans, we focus on supervised fine-tuning for instruct-following tasks, while deferring exploration of reward-based methods to future work. We explore different data combination and training settings, finding that leveraging the existing knowledge and capabilities of the foundation model with a limited dataset of instruction-focused tasks yields better results than merely increasing the volume of instruction data \cite{lima}, even when the instructional responses are of high quality. This is consistent with prevailing views that highlights the importance of building a strong foundation model. One of our best-performing instruct models, namely Tele-FLM-Chat, is demonstrated at \url{https://www.modelscope.cn/studios/FLM/ChatFLM/summary}. 

Further, our exploration into progressive learning facilitated the development of a 1T model from the initial 52B checkpoint. The central strategy involves expanding the model's structure during the pre-training phase and utilizing function-preserving growth techniques \cite{net2net, msg, flm101b} to transfer knowledge seamlessly from one stage to the next. Guided by empirical results from smaller models, we expanded the 52B model to 102B and ultimately to 1T parameters, establishing an efficient training protocol for extremely large language models without encountering post-growth divergence. We plan to release the weights of our final model, Tele-FLM-1T, to support ongoing research and facilitate further model training.

\section{Tele-FLM-Chat}
\subsection{Supervised Fine-tuning}
\paratitle{Data.} We focus on the task of Chinese instruct following. For instruct data, we collect 1 million open-sourced instructs as our full corpus, and sample different subsets from it to study the influence of various domains on model performance. The quality of responses is crucial, so we do extensive work to improve data quality.

Although the SFT data we curated covers diverge topics and intentions, and is equipped with improved responses, we observe detrimental impacts by fine-tuning Tele-FLM-52B with the entire dataset. Our investigations spanned multiple domain combinations and ratios, including chatting, mathematics, coding, reasoning, data processing, language understanding, brainstorming, and generation. Among them, our best results come from using a subset of 30k samples, including: (1) 25k samples that are categorized as ``maths'' by external clustering methods, which includes textual questions on mathematical problems (mainly in elementary and junior school level), along with general questions about mathematical concepts, and (2) 5k samples containing coding problems and multi-turn dialogues. We first sample a larger set from these domains and calculate the perplexity of the responses using our Tele-FLM-52B base model \cite{teleflm}. We leverage 50\% of the samples exhibiting the lowest perplexity for SFT.

\paratitle{Training Settings.}
We fine-tune Tele-FLM-52B for 4 epochs with a global batch size of 128. We set the learning rate to 2.7e-5, which equals to the end of the pre-training stage. The learning rate is decayed to 1e-9 with a linear schedule. The best result is achieved with the checkpoint corresponding to the end of the second epoch.

\subsection{Evaluation}
We evaluation Tele-FLM-Chat with AlignBench \cite{alignbench}, a public alignment evaluation benchmark, as well as TeleEval, an internal evaluation benchmark with a similar organization and mechanism.
\subsubsection{AlignBench}
We evaluate the alignment performance of \modelname-Chat in Chinese across various domains utilizing AlignBench \cite{alignbench}.
AlignBench is a comprehensive and multidimensional evaluation benchmark designed to assess Chinese large language models' alignment performance. It encompasses 8 categories with a total of 683 question-answer pairs, covering areas such as fundamental language ability (\textit{Fund.}), Chinese advanced understanding (\textit{Chi.}), open-ended questions (\textit{Open.}), writing ability (\textit{Writ.}), logical reasoning (\textit{Logi.}), mathematics (\textit{Math.}), task-oriented role playing (\textit{Role.}), and professional knowledge (\textit{Pro.}).
This benchmark furnishes questions, model responses, scoring criteria, and reference answers for model assessment, with GPT-4 and CritiqueLLM \cite{critiquellm} serving as judge models. These judge models provide scores and scoring rationales based on the provided criteria.

\begin{table}[htbp]
  \centering
  \caption{\textbf{Performance of Tele-FLM-Chat and baselines on Alignbench, rated
by CritiqueLLM.}}
\footnotesize
\scalebox{0.83}{
    \begin{tabular}{lcccccccccccc}
    \toprule
    \multicolumn{1}{c}{\multirow{2}[4]{*}{\textbf{Model}}} & \multirow{2}[4]{*}{\textbf{Overall}} & \multicolumn{3}{c}{\textbf{Reasoning}} &       & \multicolumn{7}{c}{\textbf{Language}} \\
\cmidrule{3-5}\cmidrule{7-13}          &       & \textbf{Avg.} & \textbf{Math.} & \textbf{Logi.} &       & \textbf{Avg.} & \textbf{Fund.} & \textbf{Chi.} & \textbf{Open.} & \textbf{Writ.} & \textbf{Role.} & \textbf{Pro.} \\
    \midrule
    gpt-4-1106-preview & 7.58  & 7.11  & 7.39  & 6.83  &       & 8.05  & 7.69  & 7.07  & 8.66  & 8.23  & 8.08  & 8.55  \\
    gpt-4-0613 & 6.83  & 6.41  & 6.49  & 6.33  &       & 7.26  & 7.16  & 6.76  & 7.26  & 7.31  & 7.48  & 7.56  \\
    chatglm-turbo & 6.36  & 4.99  & 4.88  & 5.09  &      &7.73 & 7.50  & 7.03  & 8.45  & 8.05  & 7.67  & 7.70  \\\hline
    Tele-FLM-Chat & 6.20  & 4.61  & 4.21  & 5.00  &      &7.79 & 7.22  & 7.64  & 8.53  & 8.08  & 7.72  & 7.59  \\\hline
    vs. gpt-4-1106 (\%) & 82 & 65  & 57  & 73  &     &\textbf{97} & 94 & 108  & 98  & 98  & 95  & 89   \\
    vs. gpt-4-0613 (\%) & \textbf{91} & 72  & 65  & 79  &     &\textbf{107} & 101 & 113  & 117  & 111  & 103  & 100   \\
    \bottomrule
    \end{tabular}%

    }
  \label{tab:alignbenchres}%
\end{table}%

Results on AlignBench are illustrated in Table \ref{tab:alignbenchres}. With a 52B base model and 30k SFT samples, Tele-FLM-Chat reaches 91\% of GPT-4's performance, and 82\% of the more advanced GPT-4-1106-preview. Notably, Tele-FLM-Chat is comparable (97\%) or even outperforms (107\%) the GPT-4 series on Chinese language understanding and generation tasks. However, there remains a significant gap on tasks related to math and reasoning. Remind that a large portion of our SFT data was concentrated on mathematics, with very few samples regarding other tasks are included. Thus, the results indicate that a strong foundation model requires only a few samples to elicit its capability to follow instructs well in routine tasks, as supported by existing research \cite{lima}; however, for more complicated tasks like maths and reasoning, 30k data might not be enough, and supervision on intermediate steps and chain of thoughts may be necessary \cite{verify}.

\subsubsection{TeleEval}
\begin{table}[h]
\centering
\caption{\textbf{Performance of \modelname-Chat vs. GPT-4-Turbo on TeleEval benchmark.}}
\footnotesize
\scalebox{0.77}
{
\begin{tabular}{cc|ccccccc}
\toprule
\textbf{Model}  & \textbf{Average} & \textbf{Chat}     & \textbf{Pro.} & \textbf{Trans.} & \textbf{Logic} & \textbf{Writing} & \textbf{Truth.} & \textbf{Safety} \\\midrule
GPT-4-1106-preview  &   90.7 & 92.0     & 93.8 &   97.7  & 88.4& 91.7 & 88.7 & 83.0 \\\hline
\modelname-Chat &   85.0  &90.7     &  90.3  & 93.0   & 61.0 &91.7 & 88.4 & 79.7\\\hline
vs. GPT-4-1106 (\%) & 93 & 99 & 96 & 95 & 69 & 100 & 99 & 96 \\
\bottomrule
\end{tabular}
}
\label{tab:eval_dx_new}
\end{table}

Drawing on experience from various established evaluation datasets, we develop our own evaluation dataset, namely TeleEval, which places a greater emphasis on aspects such as mathematics, security, and anti-hallucination. TeleEval includes a subset of 2500 single-turn instruct-following tests, which are categorized into 7 domains: daily chat and question answering (\textit{Chat}), professional question answering (\textit{Pro.}), translation (\textit{Trans.}), logical thinking (\textit{Logic}), long article generation (\textit{Writing}), truthfulness and anti-hallucination (\textit{Truth.}), and security test (\textit{Safety}). These test sets are distinct from, or only marginally similar to, existing benchmarks.

Results on TeleEval are presented in Table \ref{tab:eval_dx_new}. Consistent with findings from AlignBench, our results demonstrate that a small amount of maths, code, and multi-turn dialog instructions surprisingly yields a decent SFT model on almost all domains. However, maths and reasoning tasks remain as exception. Notably, on TeleEval, Tele-FLM-Chat reaches 93\% of GPT-4-1106-preview's performance.

\section{Tele-FLM-1T}

Similar to our previous work of FLM-101B \cite{flm101b}, the training of Tele-FLM-1T utilizes a staged growth strategy. This technique is essential for training large-scaled models with extremely limited computational budgets.

\subsection{Model Architecture}

Based on growth technology, the Tele-FLM-1T model training is divided into three stages by parameter size: 52B, 102B, and 1TB. Tele-FLM-52B \cite{teleflm} represents the intermediate result of the 52B stage. Each stage of the model uses the same backbone structure.
Tele-FLM models utilize the standard GPT-style decoder-only transformer architecture \cite{radford2019language} with pre-normalization and an added LayerNorm to the output of the last layer. RMSNorm \cite{RMSNorm} is used for normalization, and SwiGLU \cite{swiglu} for the activation function. Rotary Positional Embedding (RoPE) \cite{rope} is employed. The embedding layer is untied from the language modeling head. Linear bias disabled in the attention and all MLP modules.

Table~\ref{tab:model_architecture_parameters} details the architecture of all Tele-FLM models. The models grow in two phases: 52B to 102B, and 102B to 1T. The \textit{layer\_num} increases from 64 to 140, and the \textit{hidden\_dim} from 8,192 to 20,480.
In the SwiGLU setup, \textit{ffn\_dim} defaults to $hidden\_dim \times 8/3$, which is 21,824 for the 52B model, and 27,264 for 102B, respectively. In the 1T stage, \textit{ffn\_dim} increases to 98,304 (default is 54,592) to improve knowledge representation and efficiency. The \textit{vocab\_size} is 80,000. The parameter count for the three stages are 52.85B, 102.3B, and 1083.74B, respectively.

\begin{table}
  \centering
  \caption{\textbf{Detailed model architecture of each stage.} \textit{hidden\_dim} is the size of feature embedding; \textit{ffn\_dim} is the intermediate layer size of the FeedForward Networks (FFN); \textit{head\_num} is the number of heads in each Multi-Head Attention (MHA) module; \textit{layer\_num} is the total number of Transformer blocks; \textit{vocab\_size} is the size of the vocabulary space.}
  \scalebox{0.7}
  {
    \begin{tabular}{l|cccccc}
    \toprule
    Models & \multicolumn{1}{l}{\textit{layer\_num}} & \multicolumn{1}{l}{\textit{head\_num}} & \multicolumn{1}{l}{\textit{hidden\_dim}} & \multicolumn{1}{l}{\textit{ffn\_dim}} & \multicolumn{1}{l}{\textit{vocab\_size}} & \multicolumn{1}{l}{Params Count} \\
    \midrule
    Tele-FLM-52B   & 64    & 64    & 8,192  & 21,824 & 80,000  & 52.85 B \\
    Tele-FLM-102B  & 80    & 80    & 10,240 & 27,264 & 80,000  & 102.3 B \\
    Tele-FLM-1T    & 140   & 160   & 20,480 & 98,304 & 80,000  & 1083.74 B \\
    \bottomrule
    \end{tabular}%
    }
  \label{tab:model_architecture_parameters}%
\end{table}%


\subsection{Growth Strategies}

In the Tele-FLM-1T training protocol, we implement aggressive growth with an enhanced growth strategy originating from our previous work MSG~\cite{msg}, a methodology that achieves strict function-preserving growth. Building on this, we further refine the depth expansion technique and explore hyperparameter tuning methods applicable to the MSG scheme.

\paratitle{Width Growth.} Under the MSG framework~\cite{msg}, width growth refers to increasing the hyperparameters regarding \textit{hidden\_dim}, \textit{head\_num}, and \textit{ffn\_dim}. In this work, \textit{kv\_channels} (dimension of the projections in MHA) is maintained at 128, establishing a relationship where $head\_num=hidden\_dim / kv\_channels$. Therefore, we only utilize the growth of \textit{hidden\_dim} and \textit{ffn\_dim}. 

Typically, \textit{ffn\_dim} and \textit{hidden\_dim} have a default ratio. In a traditional Transformer, \textit{ffn\_dim} equals to $\textit{hidden\_dim} \times 4$, while with SwiGLU, \textit{ffn\_dim} equals to $\textit{hidden\_dim} \times 8/3$. However, our previous experiments indicate that this default ratio is not always optimal. Therefore, we search for the optimal \textit{ffn\_dim} values at each growth stage.

The main idea of MSG is to use external masks to neutralize the effects of new structures on the model's function. Initially, with the mask set to 0, the function remains strictly preserved. Over time, we gradually increase the mask to integrate the influence of new structures, ultimately reaching the target structure with the mask set to 1. This integration is governed by a linear mapping function between the training steps and the mask values, utilizing a hyperparameter named as the \textit{growth\_transition\_step}, which equals to the total steps through which the mask vanishes.

\paratitle{Depth Growth.} Depth growth refers to increasing \textit{layer\_num}.  For each new layer, we initiate the parameters by duplicating those of an adjacent layer. During training, we implement a mask mechanism similar to Width Growth. When the mask is set to 0, the computation flow remains unchanged from its state prior to expansion. A similar decay scheduler based on the \textit{growth\_transition\_step} hyperparameter ensures a smooth and controlled growth transition.

Our observations indicate that copying different layers significantly affects the convergence of the post-growth model. Therefore, we propose a layer selection method based on the input-output distance of the layers. Throughout the training process, we track the changes in the Euclidean and cosine distances of hidden states between the inputs and outputs of each layer. As training progresses, layers closer to the middle exhibit smaller input-output distances, while those at the head and tail show larger distances. This is a representation collapse issue specific to the pre-normalization architecture\cite{xie2023residual}. Based on this distance metric, we derive the following layer selection criteria: (1) Prioritize layers with the smallest distance metric (we found Euclidean distances work better than commonly-used cosine distance \cite{yi9b}); (2) In cases of comparable distances, select layers nearer the end of the sequence; (3) Avoid duplicating parameters from any single layer more than twice. These guidelines ensure that the expanded model's computation flow aligns closely with that of the pre-growth state, promoting enhanced model convergence.

\subsection{Pre-training Details}


\begin{table}
  \centering
  \caption{Training Settings.}
    \begin{tabular}{lccc}
    \toprule
          & \multicolumn{1}{l}{Tele-FLM-52B} & \multicolumn{1}{l}{Tele-FLM-102B} & \multicolumn{1}{l}{Tele-FLM-1T} \\
    \midrule
    Learning Rate Begin & 1.500e-4 & 2.740e-5 & 2.740e-6 \\
    Learning Rate End & 2.781e-5 & 1.370e-6 & 1.830e-6 \\
    Matrix Learning Rate Begin & 1.500e-4 & 2.191e-5 & 2.192e-6 \\
    Matrix Learning Rate End & 2.781e-5 & 1.096e-6 & 7.321e-7 \\
    LR Schedule Type & cosine & linear & linear \\
    Warmup Step & 2000  & 2000  & 0 \\
    Batch Size (tokens) & 5.5M  & 5.5M  & 2M \\
    Consumed Tokens & 2003B & 300B  & 15.7B \\
    Input Mult & 1.0   & 1.0   & 1.0 \\
    Output Mult & 3.125e-2 & 2.500e-2 & 2.500e-2 \\
    Growth Transition Step &       & 2000  & 200 \\
    Width Growth Init Strategy &       & Normal Distribution & Normal Distribution \\
    Width Growth Init STD &       & 0.004 & 0.004 \\
    Depth Growth Strategy &       & Distance-based Rule  & Distance-based Rule \\
    \bottomrule
    \end{tabular}%
  \label{tab:training_settings}%
\end{table}%

Table~\ref{tab:training_settings} presents detailed parameter configurations for training the Tele-FLM-1T model. The entire training process utilizes 2318.7B training tokens. Consistent dataset proportions are applied across all three stages for stable convergence. Separate learning rates are set for vector-like and matrix-like parameters, with each stage's initial learning rate carefully calibrated not to exceed the order of magnitude of the final learning rate of the previous stage. Optimal values are determined through parameter search. A linear schedule ensures a quick early decrease in learning rate for the latter two stages.

In the post-growth transition phase, both the Warmup Step and the Growth Transition Step are crucial for stability. An insufficient number of steps can lead to short-term loss fluctuations, while an excessive number may retard the descent of loss. These values should be adjusted in conjunction with the learning rate. 

For initialization, we use a normal distribution with a standard deviation (STD) of 0.004 for newly expanded parameters in Width Growth. In Depth Growth, we employ the layer selection rules based on input-output distances as discussed earlier.

\section{Lessons Learned}
We summarize the lessons learned from our explorations in supervised fine-tuning and progressive growth, with our Tele-FLM (FLM-2) model series.

\paratitle{Supervised Fine-tuning.}
\begin{itemize}
    \item We observe a successful case following the philosophy of ``less is more'' for fine-tuning. We find that learning the instruct-following format from a modest amount of data is sufficient for a wide range of language understanding and generation tasks. This indicates that SFT has the capability to elicit the  the latent knowledge embedded within pre-trained models.
    \item For reasoning tasks (maths, logical reasoning, coding), more sophisticated techniques and larger volumes of high-quality data may be required, as we observe that fine-tuning with elementary-level maths problems does not translate well to more challenging tasks.
\end{itemize}

\paratitle{Progressive Growth.}
\begin{itemize}
    \item Function-preserving growth proved viable in training models exceeding 100B parameters, and capable to handle the rapid growth from 102B to 1T. In the whole training process, we observe that the model manages to recover the knowledge learned in the previous stage. It is critical to preserve a proper aspect ratio to ensure convergence and loss reduction. We also observe that each increase in model size resulted in a more pronounced rate of loss reduction. However, since training a 1T model from scratch requires immense resources, we can not investigate the actual performance of our 1T model vs. one trained from scratch, nor could we conduct a comprehensive benchmark evaluation for the 1T model due to limited computational budgets.
    \item As the model scale exceeds 1T, the optimization issue becomes intricate, especially when the model is not trained from scratch, but grown from an existing one. Some experiences and inspirations from small model training are beneficial for the large model scenarios, while others are not. Our initial observations indicate that the training hyperparameters listed in Table \ref{tab:training_settings} are effective, yet further exploration is necessary to refine the operators, initialization, and schedules.
\end{itemize}

\section*{Acknowledgments}

This work is supported by the National Science and Technology Major Project (No. 2022ZD0116300) and the National Science Foundation of China (No. 62106249).
We would like to thank Boya Wu, Li Du, Quanyue Ma, Hanyu Zhao, Shiyu Wu and Kaipeng Jia for their help on data, Hailong Qian, Jinglong Li, Taojia Liu, Junjie Wang, Yuanlin Cai, Jiahao Guo, Quan Zhao, Xuwei Yang, Hanxiao Qu, Yan Tian and Kailong Xie for their help on computational resources, and all other colleagues' strong support for this project.

\bibliographystyle{plain}
\bibliography{custom}

\end{document}